\documentclass{article} 
\usepackage[small]{caption}
\usepackage{graphicx}

\usepackage[american]{babel}

\usepackage{natbib} 
\usepackage{mathtools} 
\usepackage{booktabs} 
\usepackage{tikz} 

\usepackage{subfigure}
\usepackage{float}
\usepackage{hyperref}
\usepackage{color}

\title{Purposeful and Operation-based Cognitive System for AGI}

\begin{document}
\maketitle

\begin{abstract}
  This paper proposes a new cognitive model, acting as the main component of an AGI agent. The model is introduced in its mature state, and as an extension of previous models, DENN, and especially AKREM, by including operational models (frames/classes) and will. In addition, it is mainly based on the duality principle in every known intelligent aspect, such as exhibiting both top-down and bottom-up model learning, 
  generalization verse specialization
  , and more.
  Furthermore, a holistic approach is advocated for AGI designing and cognition under constraints or efficiency is proposed, in the form of reusability and simplicity.
  Finally, reaching this mature state is described via a cognitive evolution from infancy to adulthood, utilizing a consolidation principle.
  The final product of this cognitive model is a dynamic operational memory of models and instances.
\end{abstract}

\section{Introduction}

Our consistent goal is to construct a basic realistic model for \textit{AGI} (Artificial General Intelligence). It is a gradual process with many versions along the way.
Hence, this paper presents \textit{MOM} (Model Of Models), the next version of \textit{AKREM} (Associative Knowledge Representation) \citep{10.1007/978-3-031-19907-3_6}. 

\textit{AKREM} is a mature-state knowledge representation model, based mainly on the assumption that communication is about encoding the sender's will into a sequence of words (a message), and then decoding it by the recipient. The model proposes a representation of any message, in a hierarchical form based on grouping, by generating some essence in a given level, from details in the lower level. The lowest level details are founded upon some DNN (Deep Neural Network), generating the basic concepts and actions (from which the details are made of) from unstructured input.
Finally, while the will concept exists in \textit{AKREM}, it will be expanded upon in this paper. 

Following this, new additions to the presented model, including new associations, operationability, modeling, consolidation, and reusability, are introduced.
First, while \textit{AKREM} assumes that the learned \textit{elements} are either objects or static actions (verbs), new associations are introduced: object's attributes and relations. Next, these connections are all static representations of knowledge, i.e., the hierarchies cannot be changed.
Therefore, operationability introduces a new type of association to objects: actions that act upon them, thus producing new knowledge \textit{elements}. This makes the connections in \textit{AKREM}'s hierarchies dynamic, hence it allows the freedom to update and create new hierarchies.
Next, modeling introduces some basic cognitive operations, e.g. abstraction and grouping. Both gather many details into fewer. Grouping specifically is about connecting \textit{elements} via some common property. It could be for example a chronology in a plot or other common properties/actions grouped into classes.
Finally, consolidation is a process in time, that collapses a huge amount of possibilities into small set of patterns, of any kind.

All the operations above are considered to be bidirectional, i.e. everything lies in some range between extremes, i.e. everything has its inverse (dualism). In grouping, it is from the whole to its parts and vice versa, and in abstraction, it is from instances to classes and vice versa. Consolidation is an operation in time, creating models and memory, while its inverse is forgetting, which is also an operation in time.
Will also lies in the dichotomy of determinism and randomness.

The product of these cognitive operations is a dynamic memory of models, formed as a semantic network of \textit{elements}. It encourages a holistic approach for AGI designing: one simple system for multiple functions, such as short-term and long-term memories, problem-solving, communication, learning, and any cognitive function. Moreover, if in the early epoch of AI, symbolic reasoning was dominant, and nowadays connectionism dominates, then we come to a new era, where we should combine and include many conflict perspectives, in cooperation and competition. Our holistic perspective embrace this duality and other dualities also. 

Lastly, operational modeling presents \textit{AGI} agent's intelligence in its mature state, which is the state of how its knowledge should be represented. However, to accomplish this state, a cognitive evolution over time is required, which utilizes the consolidation principle.

\section{Will} \label{sec:AKREM}

In this section, the importance of will as an essential element in human intelligence is elaborated upon, starting from the previously presented model, \textit{AKREM}.

Will in \textit{AKREM} is represented in the levels of any specific hierarchy. 
Starting from the most detailed aspects of will, at the lowest level, and finishing at the most abstract will or its essence, at the top. The top level represents some kind of experience uniqueness to differentiate it from other memories that use the same low-level structures.

This hierarchical will is especially demonstrated in a constrained environment, as our reality, for topics like problem-solving and communication. 
In problem-solving, the main will produces sub-wills in lower levels, till it reaches the final solution at the bottom level (Fig~\ref{fig:will_in_constraint_environment}(a)). The final result is a plan or a sequence of actions. See more in Appendix~\ref{sec:problem_solving}.
Similarly, in communication, the sender encodes/converts his will into a sequence of actions (in a language form), while the recipient on the other side decodes the intention/will from this sequence (Fig~\ref{fig:will_in_constraint_environment}(b)).
In both cases, evaluation is necessary, hence this top-down process is cyclic and non-linear.

\begin{figure}[H]%
	\centering
	\subfigure[3-level problem-solving in state space]{%
		\label{fig:a}%
		\includegraphics[width=0.97\textwidth]{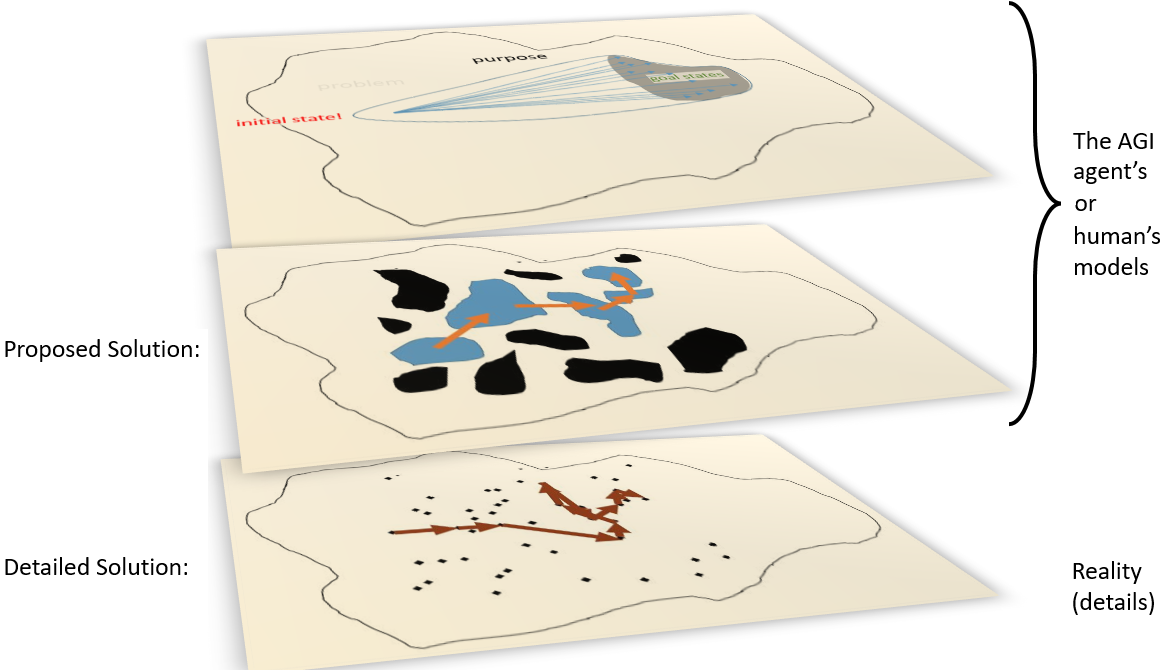}}%
	\hfill
	\subfigure[Communication model]{%
		\label{fig:b}%
		\includegraphics[width=0.94\textwidth]{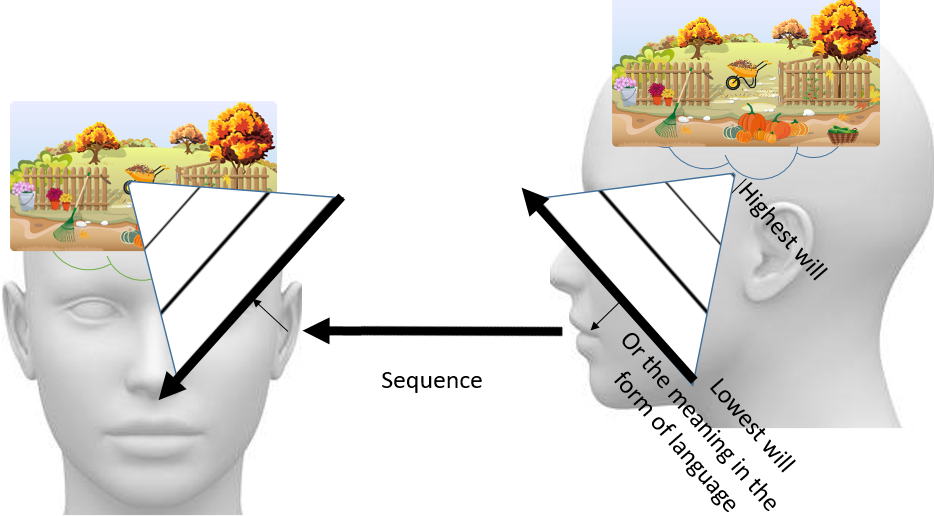}}%
	\caption{Two cases of will in a constraint environment}
	\label{fig:will_in_constraint_environment}
\end{figure}

All the above are specific cases of will, but there is also a more general will.
As recipients of reality, our main cognitive will is to find the most appropriate/simple model to fit all the pieces/details in the right place or to make the most sense of them\footnote{Making sense is also important for explainability, especially storytelling modeling can provide this. Hence, explainability should be generative and flexible, in its most general conception}. This is similar to decoding a will from a message/mystery/riddle, and it can be rephrased as a general problem-solving task to comprehend reality. First, it is done internally, by reorganizing our models (mostly during a sleep phase), and later it is done externally, in any kind of problem-solving, or in understanding a message/story/riddle/situation/phenomena.
The best model will allow us to move from place to place in it easily, perform new actions, and produce conclusions/solutions easily.
This results with an understanding, or the ability to control any aspect in the complex model. So in a sense, we have two wills governing our cognition: controlling and subsequently making sense. Obviously, these wills are enforcing each other.

\begin{figure}[H]
	\centering
	\includegraphics[width=0.95\textwidth]{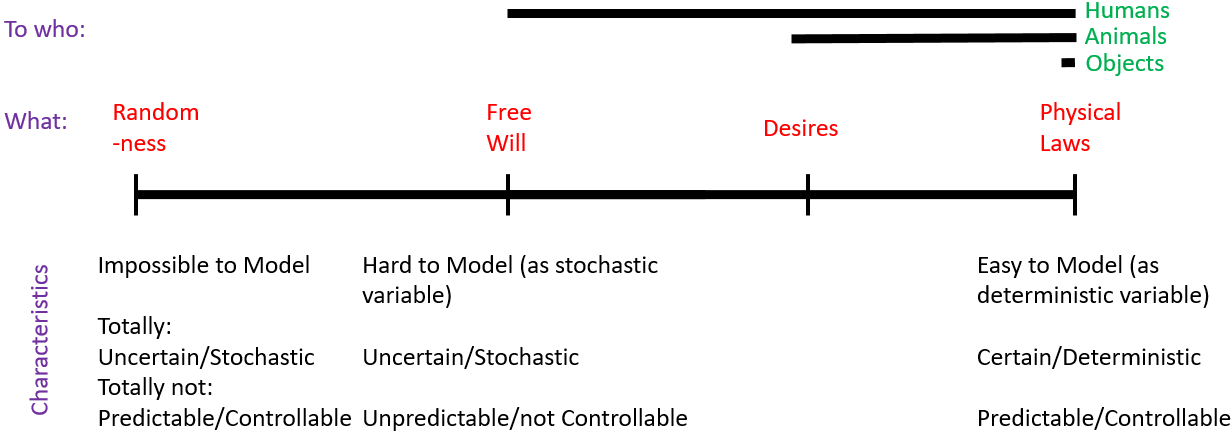}
	\caption{Types of will}
	\label{fig:will_types}
\end{figure}

Additionally, there are different categories of will, such as chronology, causation, and purposefulness. In stories, they are very intertwined/mixed. It is because purposefulness is a higher manifestation of will (usually applied in humans), while causation is a lower one, usually applied to animals/objects (e.g. "A causes B"), and chronology is simply the way will is implemented: in a delay. You first want, and then you try to accomplish it. Or in the case of causation, there is a law, as a fixed kind of will (e.g. gravity), and then it is realized. 
See more in Fig~\ref{fig:will_types}.

In conclusion, purpose is everywhere, in all of our daily interactions. We try to figure out animals' intentions or people's hidden will, to make sense, or in our case to make our model complete, i.e. enable prediction. Since a specific human has his own will, he tries to figure out other people's will to enforce its  own will over theirs. Sometimes it is via competition, while sometimes it is through cooperation.

However, how is will actually included in \textit{MOM}? It is discussed in \ref{sec:Modeling} and Appendix~\ref{sec:attention}.

\section{New Associations}  \label{sec:Associations}

Inspired by semantic nets, some general connections are proposed, such as inheritance (is-a relation), instance property (is-instance-of relation), part-whole property (part-of relation), an attribute of an object property (has relation), assigning a value to attribute (value relation), synonyms, antonyms/opposites and more, see \citep{Wiktionary}. All these connections are static and can represent stories as separate items as it is in \textit{AKREM}. But to connect these details, as a sequence of applied actions, operationability is introduced.

\section{Operationability}  \label{sec:Operationability}

It is hypothesized that thinking is operational. Meaning it is a process, generating new facts along the way, via some set of actions.
Hence, the first principle added to \textit{AKREM} is operationability, which turns it from static to dynamic knowledge representation. That is, unlike static connections as in \textit{AKREM} and knowledge/scene-graphs (representing facts or a scene shot) - action connections in \textit{MOM} can also be productive (produce new \textit{elements}).
It adds degrees of freedom to the current cognitive model, to move in new directions along the hierarchy, i.e. to create new hierarchies on the fly or update old ones, via admissible actions.

Subsequently, a minimal set of primitive operations is proposed, to function as basic operations, which can be the building blocks for more complex and composite operations/actions. 
Operations such as logical relations (AND, OR, NOT, all, each, (in)equalities, exists, count), flow operations like loop operations (while, for) and if-else conditionals, mathematical operations (+,-,*,/,min,max,norm,log), and other relations.
This set of tools can replace DNN units and DNN's fixed structure, in a program-search process. It can be implemented for example via Reservoir network, a random mix of basic rule components/blocks, yielding an algorithm best describing an operation.

At the same time, prior knowledge is needed to be inserted, within all \textit{elements}, by including: number of visits, uncertainty, rate of update, and measure of consolidation. The measure of consolidation is to prioritize different options associated to some \textit{element}, to separate the relevant from the irrelevant, e.g. admissible actions, which can be used inversely in creativity mode - by picking the less expected directions to follow.

Moreover, action's admissibility is needed for two reasons. Firstly, due to the elimination of entry conditions necessary for an action to be performed, e.g. on which types (integer, string, etc). And secondly, it is due to the ability to use 
High-Order Logic, as in $\lambda$-calculus, which removes any restrictions on an object's slot or action's argument. Hence, relevancy is needed to constrain action's admissible space.

\section{Modeling} \label{sec:Modeling}

If operationability is considered, as the addition of freedom to move in a 2D knowledge representation, then modeling is an addition of a new dimension, i.e. converting it to 3D.
It is about extending the "is-a" operation into a programming abstraction, as in \textit{OOP} (Object-Oriented Programming), or in abstract mathematics, such as algebra or category theory\footnote{
	Group properties include identity object to use compositionality of an operation and its inverse operation. This implies duality.}.
Meaning, that while usually semantic networks represent this operation in a 2D graph, here the instances are totally separated from classes, and from classes of classes, and so on. Resulting with a multi-level of abstractions, while for simplicity two types of levels can be distinguished, in the final LTM (Long-Term Memory), see Fig.~\ref{fig:Modelling_machines2}.
In summary, at first all different associations including operationability are describing objects, and then abstraction extends objects as instances into classes, which represent models.
Consequently, unlike \textit{AKREM}, these full models are important for natural communication, e.g. for context-based conversations, where undelivered missing information (common sense) is needed to be filled.

This extension has several implications.
First, in perception from senses: from the basic recognition of instances in \textit{AKREM}, to a multi-level recognition of instances and classes.
Next, every \textit{element} is learned and can be abstracted (into class), i.e. objects, actions, relations, and attributes.
For example, action with attributes, relation with attributes such as strength, numeric attribute/values as class (integer, real), group types (sets, lists, arrays) with their group operations (slicing, union, sorting), and more.
Next, this extension enables answering the question about how will is implemented in \textit{MOM}, see \ref{sec:AKREM}.
Alternatively to \textit{AKREM}'s hierarchy by will, which it is not clear how it can be implemented, it could be generated by abstraction, while the will/intention will is serving as an additional and independent variable in the models that construct the hierarchy.
Additionally, feeling measures could be included, as influencing will.
Moreover, since there are several levels of will, correspondingly there are the main variable and secondary variables representing these wills, perhaps with different significance intensities, depending on the abstraction level.

Finally, the novelty here, is that unlike \textit{DL} which performs program-search in an un-interpretable way, here however, additional inductive bias is introduced: separating of models and performing program-search to relevant actions, in consistency with other models and actions. This makes \textit{MOM} both usable and interpretable.

\subsection{Learning the modeling} \label{sec:Learning_the_modeling}

Here, model learning mechanism is proposed, where two contrary but completing learning approaches in AI are combined \citep{day2022knowledge}: empirical, i.e. from examples (induction), and expertise (rule-based). This is a learnable symbolic manipulation, or can also be referred to as a hybrid approach or Neuro-Symbolic, see \citep{marcus2003algebraic}.
Empirical is bottom-up (from examples to rules), e.g. via observation or passive interaction. Rule-based is straight from the top, via rules in abstract language, e.g. via conversation or observation. It can then descend to examples of these rules (deduction). 

These approaches might contain models of concepts that do not belong to them both, but only to one of them. For example, concepts that are hard to define, like love, God, beauty, and tacit/unconscious knowledge like walking and breathing - all can be modeled simply by examples. Similarly are the sub-symbolic features, like audio and visual inputs - they do not have logic/linguistic/symbolic meaning, hence should be modeled by examples, as it is done in \textit{DL} nowadays. Hence, these non-symbolic concepts can be learned via usual non-interpretable \textit{DL}.
On the other hand, abstract concepts, like those in math and sciences, that appear less in the physical reality, can be learned solely in the top levels of LTM.

\begin{figure}[H]
	\centering
	\includegraphics[width=0.95\textwidth]{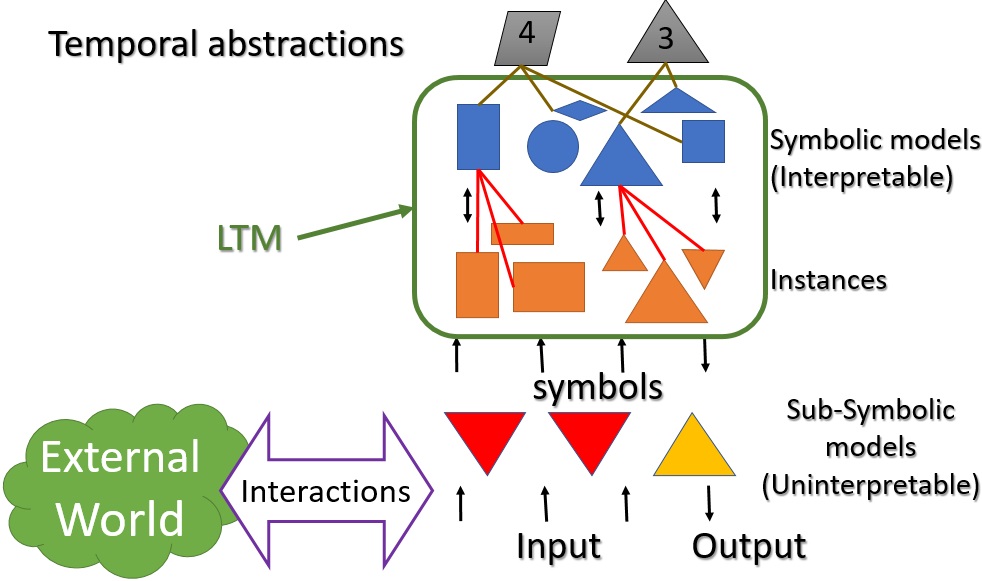}
	\caption{Proposed cognitive model basic diagram}
	\label{fig:Modelling_machines2}
\end{figure}

Moreover, in this hybrid approach, \textit{DL} (Deep Learning) is used twice. On the one hand, \textit{DL} is extended from its too constraint program-search to be much more flexible, if more operations are added as building blocks, see \ref{sec:Operationability}. Hence, symbolism is learned and adaptive just like in \textit{DL}, differently from expert/rule-based AI. On the other hand, different input sensors are fused to represent specific symbols/concepts, i.e., the uninterpreted features in \textit{DL} become symbolic tokens (Fig.~\ref{fig:Modelling_machines2}).

The reason the hybrid approach is preferred over \textit{DL}-only, is that \textit{DL} usually does not implement compositionality, modularity and abstraction \citep{Dickson}. One of the effects of this shortcoming is the shortcut effect, where the \textit{DL} categorizes something due to the wrong reasons. It can be solved in a system, that learns a model with alignment and consistency-check with other models learned so far, which \textit{MOM} supplies (continual learning).

Additionally, the topmost level is actually temporal and used for creativity and problem-solving. In this level, temporal new abstractions are created, by stripping off attributes/actions/relations, thus connecting distant or different abstractions to perform analogy or transfer learning between different domains. For example, the abstraction is via the number of edges in the polygon classes (Fig.~\ref{fig:Modelling_machines2}).

Furthermore, learning can be divided into online and offline modes. In waking periods, i.e. when sensors are active, the learning is online and minimal since most resources are dedicated to fast response (e.g. fast optimization into local optimums). However, during sleeping periods, sensors are inactive, and previous memories can be used for improving models to make overall sense, e.g. larger time scale is used to generate causal relations in models.
In such a case, it is a slow processing, implemented e.g. as a 
Neural Architecture Search or a Genetic Algorithm, to get out of local minima and search for a global one.

Finally, another issue precedes learning: how to obtain separate models at all. One way, is like the \textit{DENN} (Dynamic and evolving NN) idea \citep{10.1007/978-3-031-19907-3_5}
, i.e. always learning the "model of everything" while refining it more and more with every new experience, such that new sub-models are produced. 
The idea here, is like Jeff's hierarchy \citep{hawkins2007intelligence}, where the top model is always reached, at perception from senses, and it decides which lower model will handle the situation. For example, in recognizing a specific type of problem, it chooses the most appropriate model for solving this problem. Or, when encountering new knowledge, it selects the most appropriate model to handle its assimilation in the current memory of models.

Another way to facilitate modeling evolution is via consolidation.

\section{Consolidation}   \label{sec:consolidation}

So far, the cognitive model has been presented in its mature state. Now, the discussion is about how to reach it. This is a process in time, which is mainly based on consolidation.

Consolidation is about transforming from chaos to some stable order of patterns, or from a continuous realm to a discrete one, as in quantum mechanics. An infinite amount of details is hard to handle (i.e. to understand and then to control), therefore consolidation to fewer patterns is required.
Consolidation also allows for fuzzy logic and categories \citep{wyler1995fuzzy}.

Consolidation can be expressed in many forms, such as:
\begin{itemize}
	\item in the conversion of sub-symbolic to symbolic, for any type of \textit{element}
	\item in cognitive evolution: from flexible (at infancy) to less flexible (at adulthood)
	\item in modeling, at program search, from huge hypothesis space for possible programs to a small set of hypotheses (as in \textit{DL}). It is both in the micro (within models) and in the macro (between models)
	\item in testing multiple versions of an unknown model, and finally converging into less/one version(s) that are/is consistent with evidence
	\item and in grouping/abstraction, where some separate elements become connected
\end{itemize}

Note that causality is a special case of modeling, a spatio-temporal one, where re-occurrence is consolidated. More generally, re-occurrence help in learning both static objects and dynamic basic/composite events (equivalent to scenarios/scripts in \textit{OOP}). 

Additionally, \textit{MOM} enables multiple parallel versions of the same thing, since any specific topic or subject can have multiple theories/models, sometimes in conflict.
Hence, like in the quantum superposition realm, multiple-version combinations could be tryout, and consolidation can help in collapsing them into fewer 
versions. Those versions should make the most sense, i.e. to be consistent on different occasions or supporting the majority of evidence.
Thus, it just maybe, that at infancy, a highly uncertain period, there are many versions created, and with time - only the most consistent ones survive (consolidate).

Lastly, two operations help in producing consolidation.
On the one hand, to deal with a stochastic environment and ambiguous signals, \textbf{repetition} provides memory prioritized by relevancy. Repetition is never exactly over the same thing, but rather over many different examples of a thing. Repetition is needed also with guided tutoring of an AGI agent.
Conversely, \textbf{sparsification} is about reducing irrelevant signals.

\subsection{Reusability}

An additional form of consolidation is reusability, since the more learning progresses, the fewer new models are proposed in favor of using existing ones.
Hence, reusability is expressed via exploration (mostly at early stages) verse exploitation (in stable or mature stages), as in Reinforcement Learning. 
In the beginning, many possible codes are generated for models, but as time goes by the process is less exploratory and more exploitative, i.e. there is more emphasis on retrieving known codes, while testing fewer new codes in parallel.
In addition, reusability aligns perfectly with abstraction/grouping, in a constrained environment and limited resources. They are both needed to hold control of as much as possible, with minimum effort, i.e. without generating many models of each thing.

In practice, reusability is about using less of the initial available tools, as the learning evolves.
Meaning, while regular \textit{DL} tools (if, sum,  
activation function) or the primitive tools \ref{sec:Operationability} can be used for program search of basic action methods or relation methods, the new methods apply reusability. In such methods, less primitive tools are used while the current methods are used more, thus encouraging more connectivity in the network. 

\begin{figure}[H]
	\centering
	\includegraphics[width=0.75
	\textwidth]{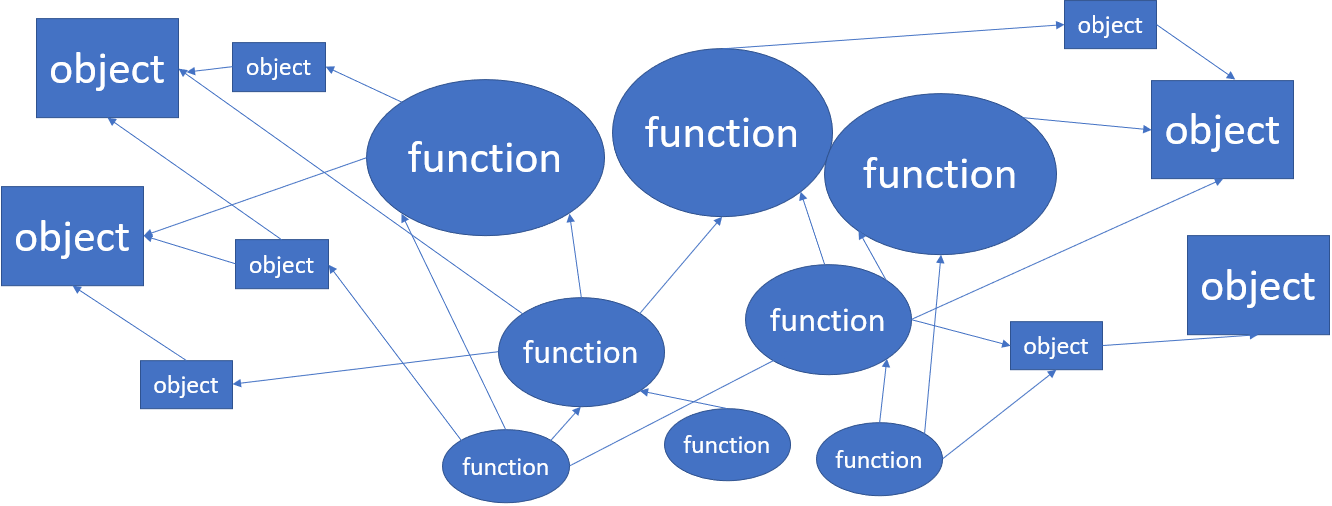}
	\caption{Actions, objects, compositionality, and reusability}
	\label{fig:COMPOSITIONALLITY}
\end{figure}

Moreover, Functional Programming can be applied to assist reusability. On the one hand, the general/outer structure is \textit{OOP}, i.e. \textit{elements} are grouped in an \textit{OOP} fashion. On the other hand, methods are kept in a pure operational immutable form \citep{Chen,van2009programming}. Meaning, having small and simple methods, which maximally reuse other functions, and without inner variables, due to objects-memory-only assumption. Meaning, methods that are comprised of other methods, as much as possible. This is compositionality/grouping applied in actions. 
See 
Fig.~\ref{fig:COMPOSITIONALLITY}.

\section{Cognitive model comparison}

\begin{figure}[H]
	\centering
	\includegraphics[width=0.95\textwidth]{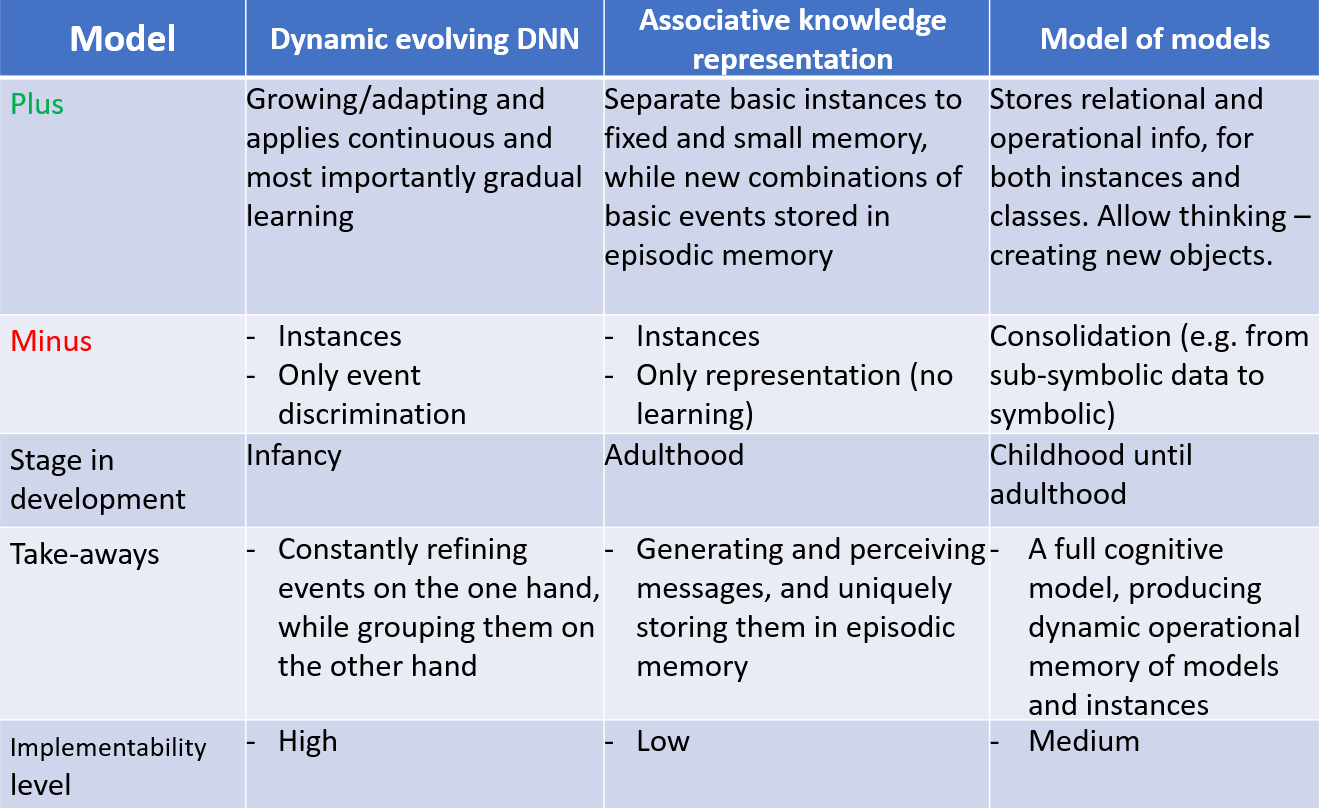}
	\caption{Comparison table of models discussed so far}
	\label{fig:comparison_table}
\end{figure}

Here all models developed so far are compared, in Fig.~\ref{fig:comparison_table}.

In summary, \textit{DENN} stores each new combination of events, while \textit{AKREM} stores dynamically in episodic memory any newly encountered combination of basic events. Events which are stored separately in two types of memory. \textit{MOM} unites those separate memories into one dynamic operational memory, consisting of concepts, actions, relations and any instance of those.

\section{Conclusion} \label{sec:Conclusion}

The following is the summary of the paper including some key takeaways.

First, a new cognitive model is introduced to join the existing cognitive architectures, in the form of dynamic operational memory of models and instances. In it, a holistic approach is embraced, assuming that intelligence should be highly versatile and diverse, instead of "picking one side".

Next, our model includes will as an essential part of the modeling process. Thus, in case of its absence, it turns most of the learned models to be very partial.
In the \textit{OOP} formulation, the will is an additional variable, and it is mostly significant in the top level while it is least significant in the lowest one.

Next, operationability turns static knowledge representation into a dynamic one, thus enabling cognitive processes. The actions are learned via regular program search, with either \textit{DL} or other tools, in a self-supervised manner.

Next, one way to ensure that the continual learning is consistent, is by implementing local learning, i.e., concentrating on updating only one/some model(s), while confirming compatibility with other models.
A model can be learned either from examples or directly by using existing operations (logically).
Another way to ensure that learning is consistent, is by a slow process of consolidation. It is ensured by maintaining a high level of flexibility over a long period, while pursuing more and more consistency within and between models.

Next, reusability is utilized to enhance connectivity between models, instead of learning them as separate entities.

Finally, the cognitive model is designed via inverse engineering. Meaning, starting from our highly aware and mature cognitive state of mind, and then tracking back in time to study its evolution.

\section{Future Work} \label{sec:future_work}
The main problem is how to implement a cognitive system, that produces the appropriate models, i.e. how grouping/clustering occur, to generate the right models. Also, how these models produce new ones by the correct compositions.

In addition, there is the issue of how sub-symbolic become symbolic. Perhaps the models produce objects and actions directly upon sub-symbolic data. Furthermore, if continuing this line of thought, then models may be removed at all, which converts this problem to pure DL-based approach.

Additionally, relevant to the last issue, is model evolving. Is it model refinement of some main model to sub-models, which controls how models of knowledge are used, or is it that all the models are separate. And if it is by refinement, is it one large DL-based model, and all the rest are knowledge models, or if we continue this line of thought - again, end up with pure DL-based one huge model, containing implicitly all the different models, their actions and attributes. But then how \textit{elements} and abstraction are implemented in such a model? More about the suggestion above see Appendix~\ref{sec:combination}.

Finally, though hierarchy by abstraction can be implemented, but how can it be implemented by will, 
i.e. how to decide when and what to group in such a hierarchy?

These are all open questions to deal with.

P.S.: Neuro-Symbolic AI for me is merely taking inspiration of class-based structure, to act as the final stage of learning, while DL is the main tool to reach it.
So it is all about flexibility of DNNs, only that we use consolidation to finally reach symbols.
Another implication of this, is memory. First, since it is vague and mostly reconstructed. It means it is not recorded accurately. And second, since it probably used in sleeping periods similarly in a vague form.

\appendix
\section{Appendix}

\subsection{Examples of MOM in action}

\begin{figure}[H]%
	\centering
	\subfigure[Developing equations (grouped chronologically)]{%
		\label{fig:a}%
		\includegraphics[width=0.85\textwidth]{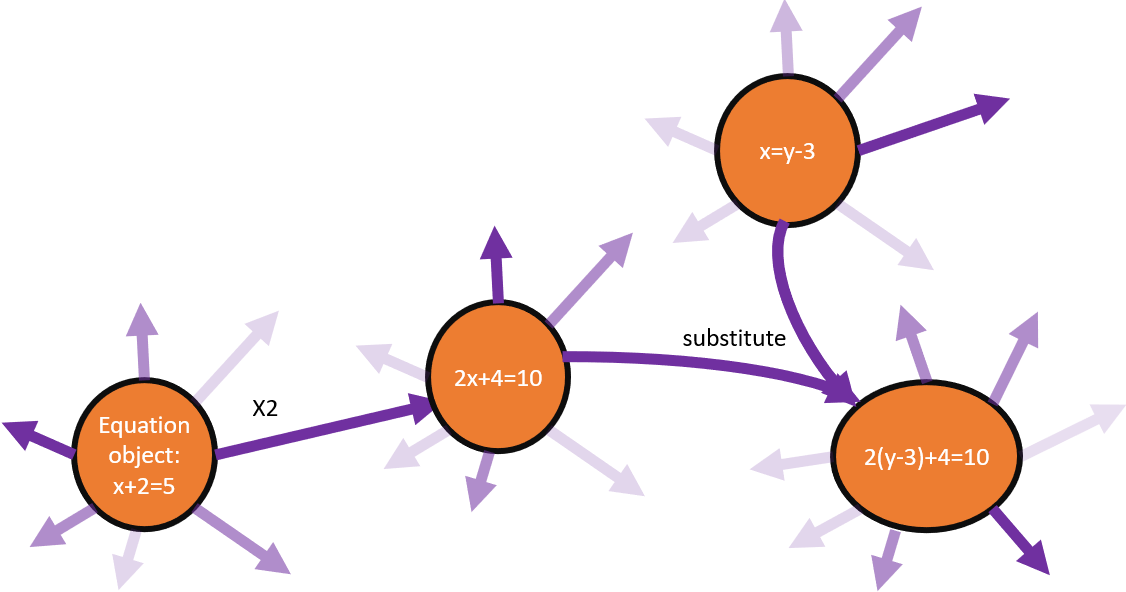}}%
	\hfill
	\subfigure[A part of a story]{%
		\label{fig:b}%
		\includegraphics[width=0.95\textwidth]{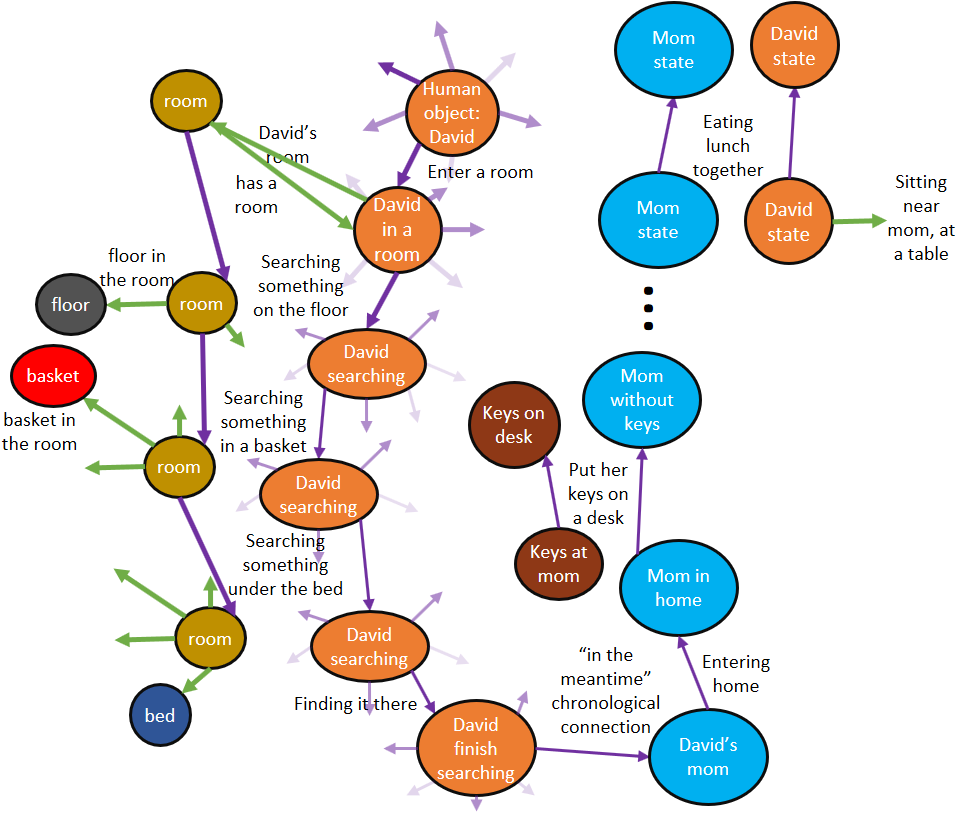}}%
	\caption{Examples of cognition processes}
	\label{fig:examples_of_cognition_processes}
\end{figure}

Fig.~\ref{fig:examples_of_cognition_processes} contains examples of cognition processes. As seen, actions denoted as arrows, can perform changes in several objects to their different attributes. High admissibility expressed via salient color. 
Fig.~\ref{fig:examples_of_cognition_processes}(b) is a \textit{MOM} representation of evolving state, representing a story. A state is consisting of several objects (joining/leaving as the story evolves), including their attributes and actions that change the state. The story: \textit{"David entered his room. He searched for something on the floor. Then he searched in the basket. Then he searched under his bed, and was thrilled to find the ball there. In the meanwhile, his mother entered home. She put her keys on the desk. Then she removed her shoes and put her sunglasses on the desk. Then she searched for David, found him, and they sat to eat lunch together"}.
The same story is represented via \textit{AKREM}, without the evolution of details, see the video link in \citep{10.1007/978-3-031-19907-3_6}. 
Note, that if repeated often, the sequence of David or anybody searching in some place can be grouped/abstracted as an event class, also referred to as \textit{Trans-Frame} \citep{minsky1988society}.

\begin{figure}[H]%
	\centering
	\subfigure[Multi-modal fusion]{%
		\label{fig:a}%
		\includegraphics[width=0.5\textwidth]{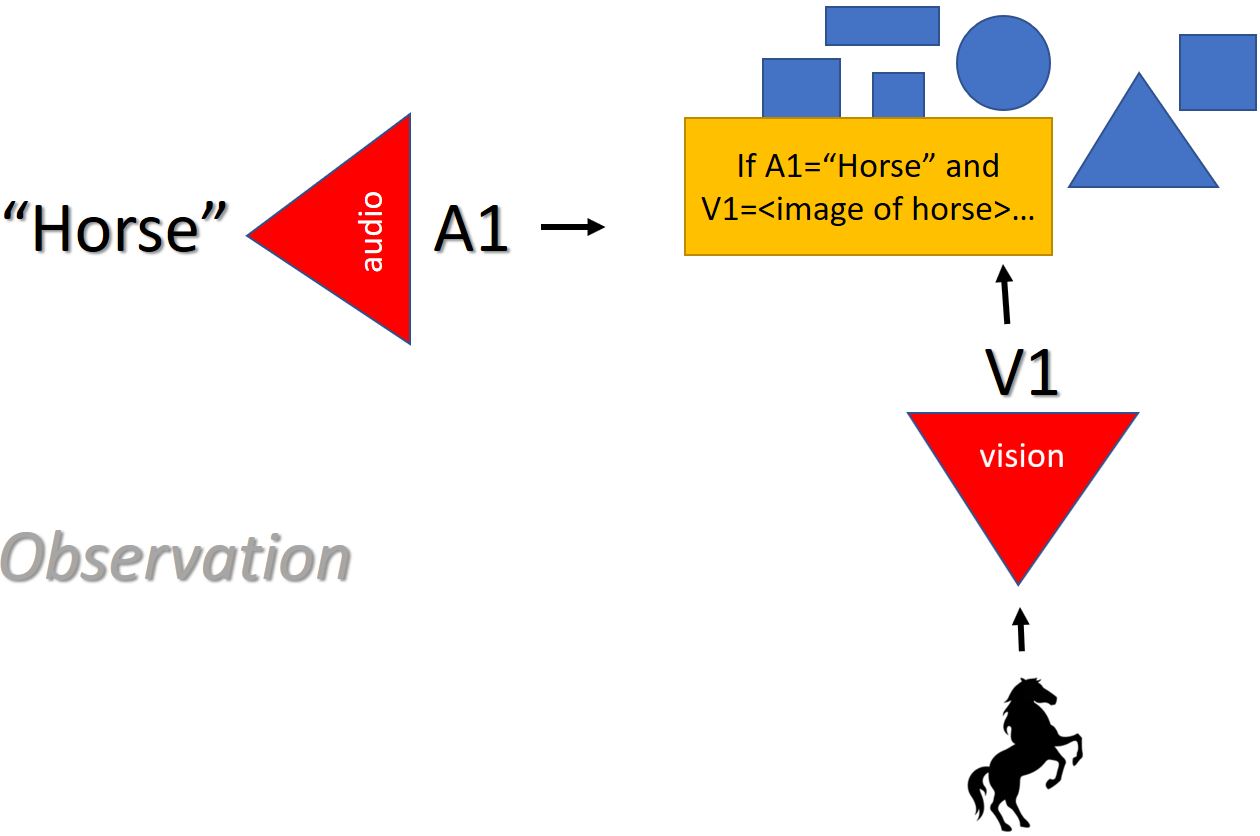}}%
	\hfill
	\subfigure[Causality relation1]{%
		\label{fig:b}%
		\includegraphics[width=0.5\textwidth]{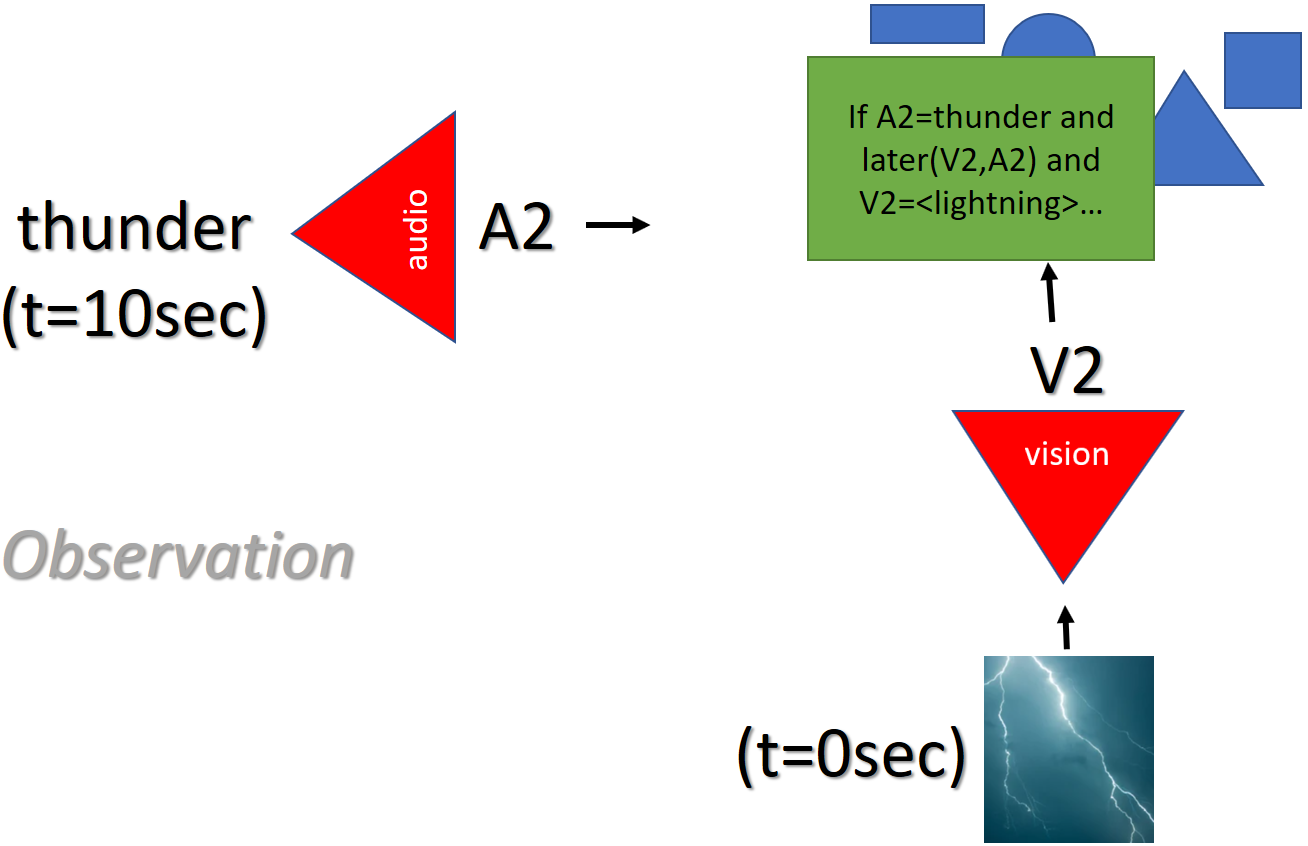}}%
	\hfill
	\subfigure[Causality relation2]{%
		\label{fig:b}%
		\includegraphics[width=0.5\textwidth]{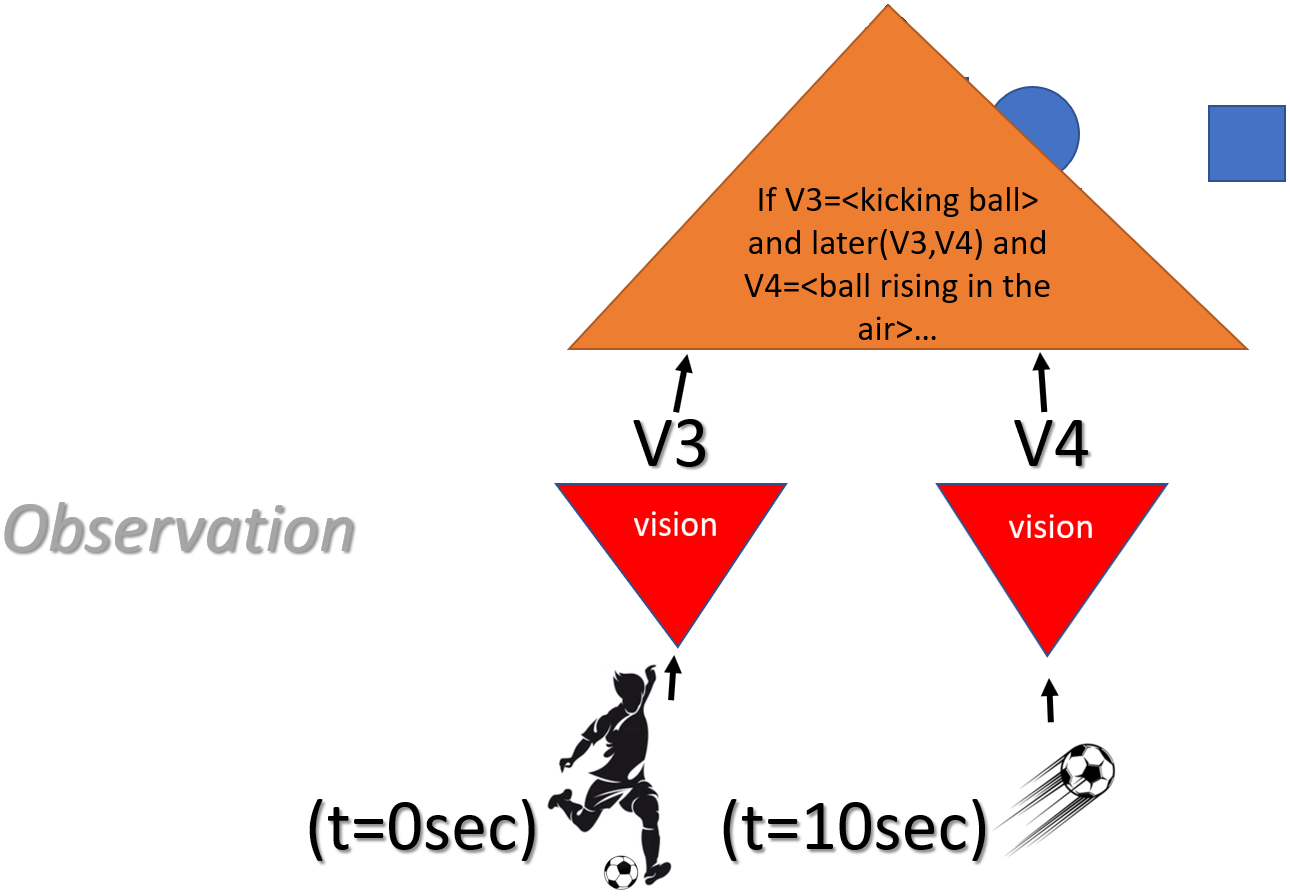}}%
	\hfill
	\subfigure[Interaction]{%
		\label{fig:b}%
		\includegraphics[width=0.5\textwidth]{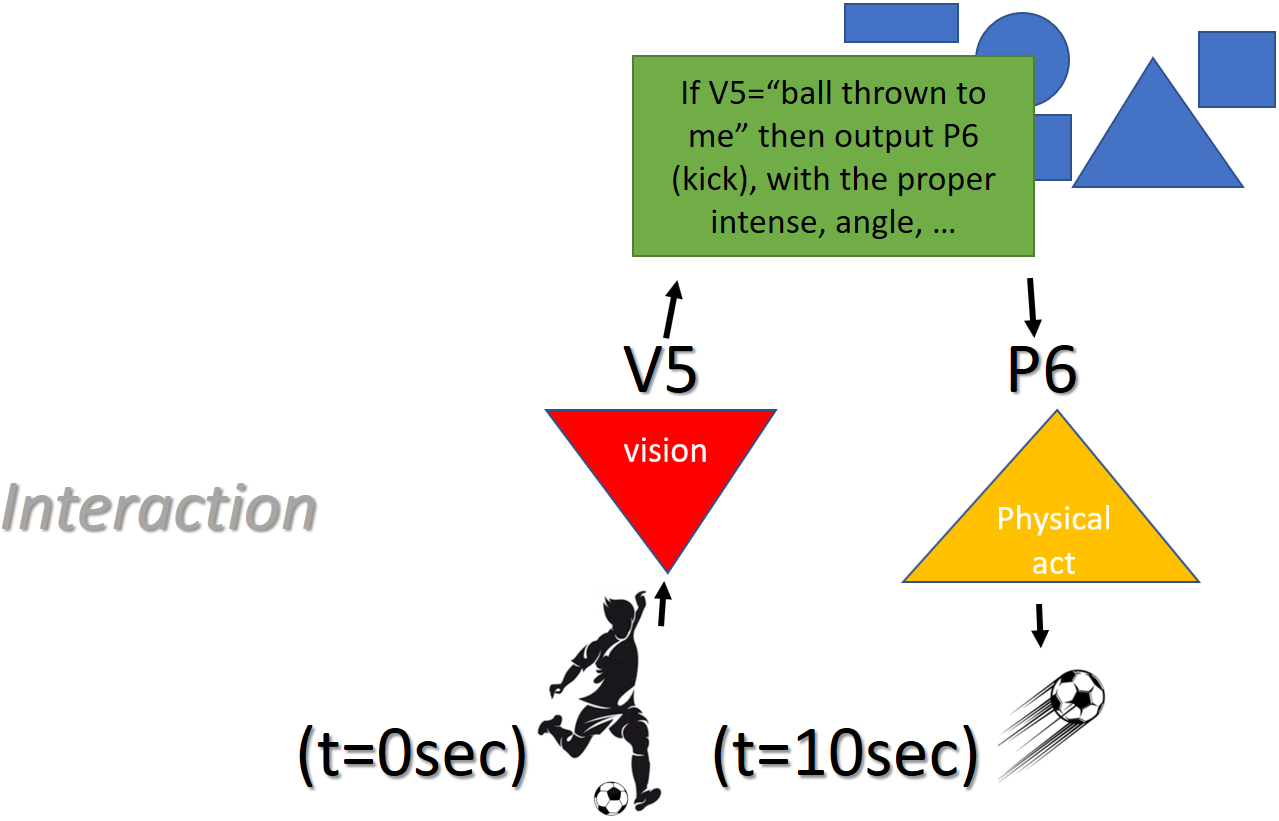}}%
	\hfill
	\textrm{\\}
	\subfigure[Reinforcement Learning (Pavlov experiment)]{%
		\label{fig:b}%
		\includegraphics[width=0.7\textwidth]{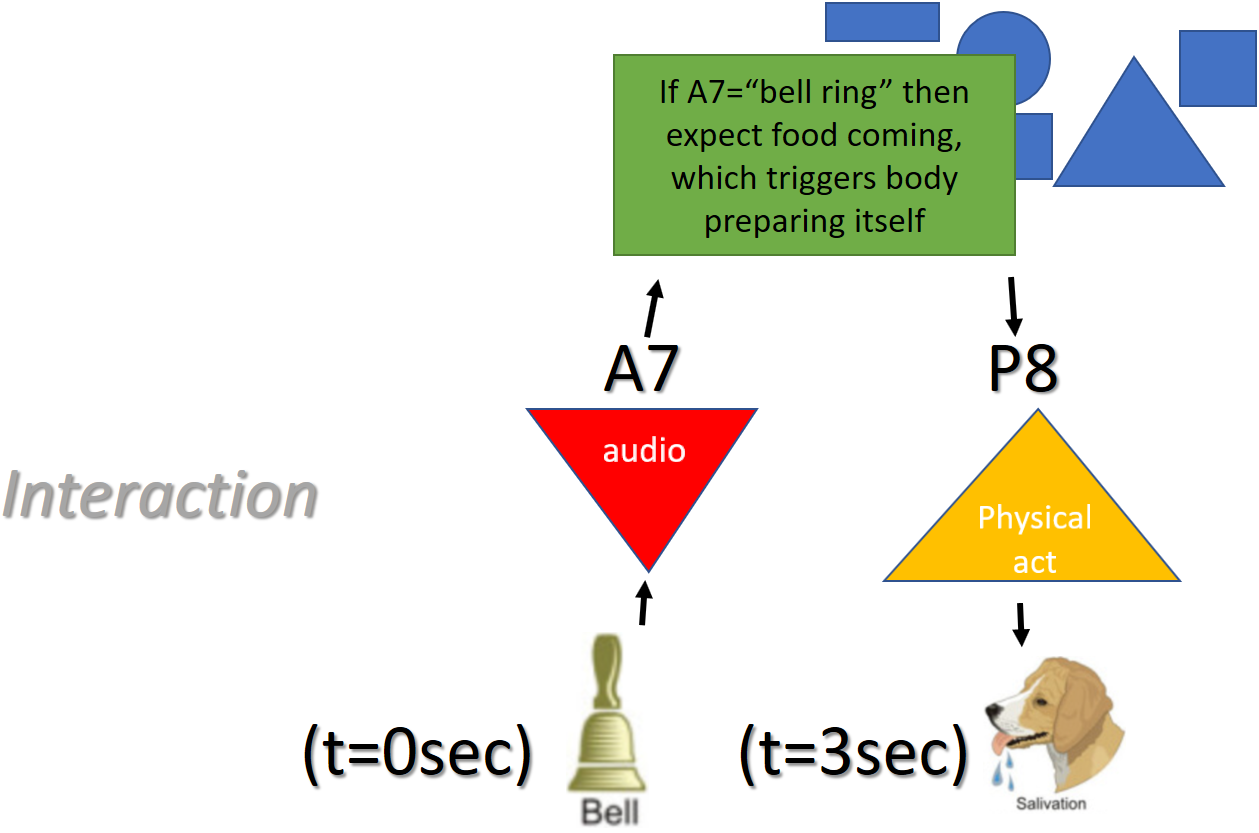}}%
	\caption{Examples of model learning}
	\label{fig:examples_of_model_learning}
\end{figure}

Next, Fig.~\ref{fig:examples_of_model_learning} contains examples of model learning, based on Fig.~\ref{fig:Modelling_machines2}. The assumption here, is that similarly to DNN pattern matching and then executing some task - here it is performed explicitly, via if condition as pattern matching, and execution following. Each sub-figure contains the type of interaction (passive or active) and the different modalities involved: A=audio, V=vision, P=physical act.

As seen, operation in \textit{MOM} involves time, i.e. it can be either immediate or include past/future of any scale, which enable causality modeling.
Also, naming, or the inclusion of language to describe objects is not necessary in model learning. The learning still occurs, even before its name is introduced, or if it is forgotten for some reason.


Finally, the conditions in  Fig.~\ref{fig:examples_of_model_learning} could be alternatively grouped/abstracted as event classes instead of being learned as an action.
Meaning, the pattern matching can be replaced by event class to be recognized, and the possible reaction to this pattern can be formed as an admissible action in such a class.
For example, "OR" can assign the same action to different objects, and "AND" assigns an action to a group of objects.

\subsection{Problem-solving and Designing Appendix}  \label{sec:problem_solving}

\subsubsection{Problem-solving}
Problem-solving is a broad topic, which is about handling any given situation, and not only solving puzzles/mysteries/science. In any such situation, we can either recognize a previous similar pattern (System 1), and apply automatic reaction, i.e. immediate resolution, or if it is not the case, try to generate a new solution (System 2). In this section, the latter option is discussed.

\begin{figure}[H]%
	\centering
	\subfigure[First: the problem]{%
		\label{fig:a}%
		\includegraphics[width=0.48\textwidth]{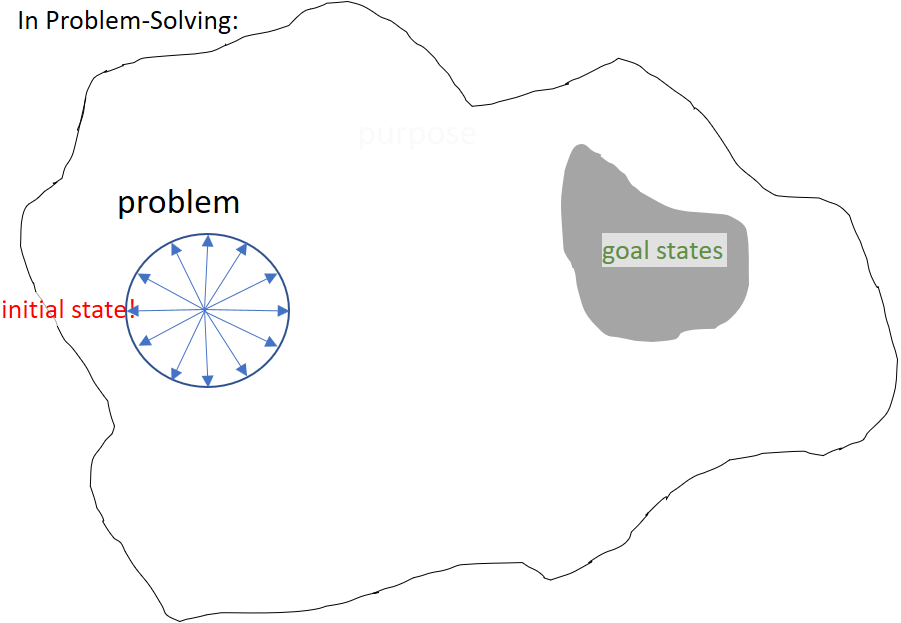}}%
	\hfill
	\subfigure[Next: will turns into a purpose]{%
		\label{fig:b}%
		\includegraphics[width=0.48\textwidth]{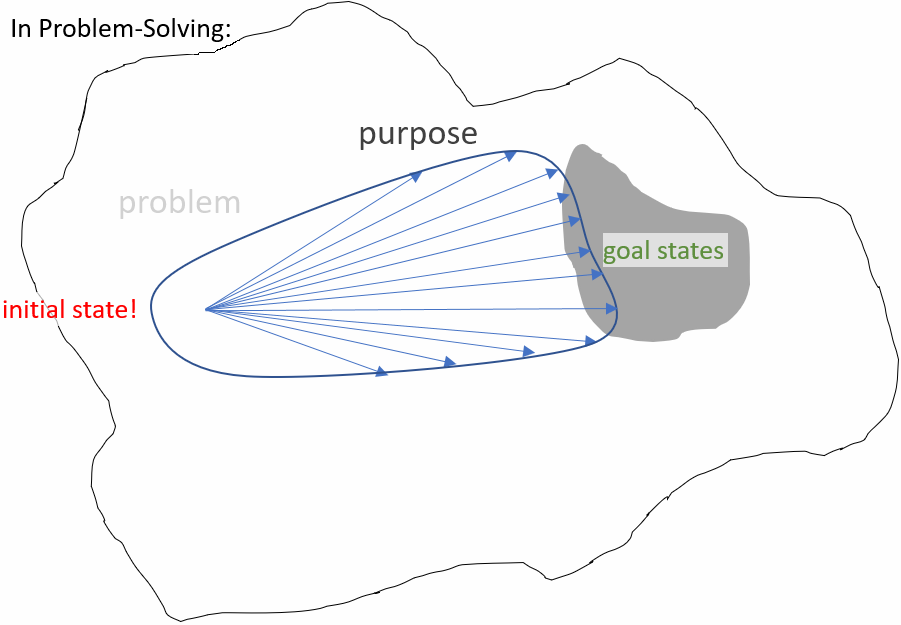}}%
	\caption{Phases of will refinement in problem-solving}
	\label{fig:Phases_in_problem_solving}
\end{figure}

In this context, it is represented within some state space, where a problem is situated at some point or a region in the space. Additionally, a problem is expressing the current (problematic) situation, involving a general will to get out of this situation. Hence, a will is not yet formulated at this stage (Fig.~\ref{fig:Phases_in_problem_solving}(a)).
Next, it is about deciding upon some goal states to be reached, to gain a desired resolution. When goal states are defined, the will become purposeful.
Purpose is a more definite will, because it gives some “direction”, either a vague direction or a strong one, to specific goal states (Fig.~\ref{fig:Phases_in_problem_solving}(b)).
Since will derives action(s), it is represented similarly to an action in the state space - as a vector, transmitting one situation to another. Meaning, will is defining the direction the agent wants to move, before it found the admissible/legitimate/allowable way to realize it, in the given environment.
Finally, the agent starts to plan how to solve the given problem,  under given constraints, i.e. where one cannot fulfill its will directly, but instead look for some legitimate way to accomplish this, in the given circumstances.

After the will is refined and transferred to a purpose, the search for a solution is initiated, and it is depicted in Fig~\ref{fig:will_in_constraint_environment}(a). This is the next phase of problem-solving: realization.
The figure demonstrates a coarse-to-fine hierarchy, where the will along with its refinement, is placed at a top level. This level is vague, since nothing is perceived clearly about the ground level. However, descending the levels reveal more and more details, and get the will more closely to realization. It is similar to the process of zooming in on a geographical map.
Higher levels propose general models as potential stations in a possible trajectory from a problem state to a goal state. Then at descending, finer models are proposed, consistent with the upper levels, to move from a problem to a goal state.
The search at any level can be performed by any heuristic/learned model, such as back/forward chaining, Depth-First Search/Breadth-First Search techniques, or any combination of those.
Note the mismatch between our model level (proposed solution) and data level (detailed solution), in Fig~\ref{fig:will_in_constraint_environment}(a). It is due to our inclination towards abstracting, i.e. memorizing the essences and less the details, which is essential for efficient learning, as described also in \ref{sec:consolidation}, where it is better to learn several patterns than to lose yourself in a non-pattern realm, where all we see are details.

In summary, this approach is non-local, i.e. similar to Means-Ends Analysis, it is looking simultaneously at the whole region, only within different resolutions. It is also cyclic and non-linear, both in the will-refining stage and in the realization stage. At will refining, it is since sometimes the goal states cannot be reached, so other states are needed to be generated, sometimes as a compromise. At the realization stage, it is since descending in levels might result in conflicts or failures, due to misalignment between the lowest models of reality and the actual reality. Hence, returning to higher levels for trying different solutions is needed.

\subsubsection{Designing}

\begin{figure}[H]
	\centering
	\includegraphics[width=0.94\textwidth]{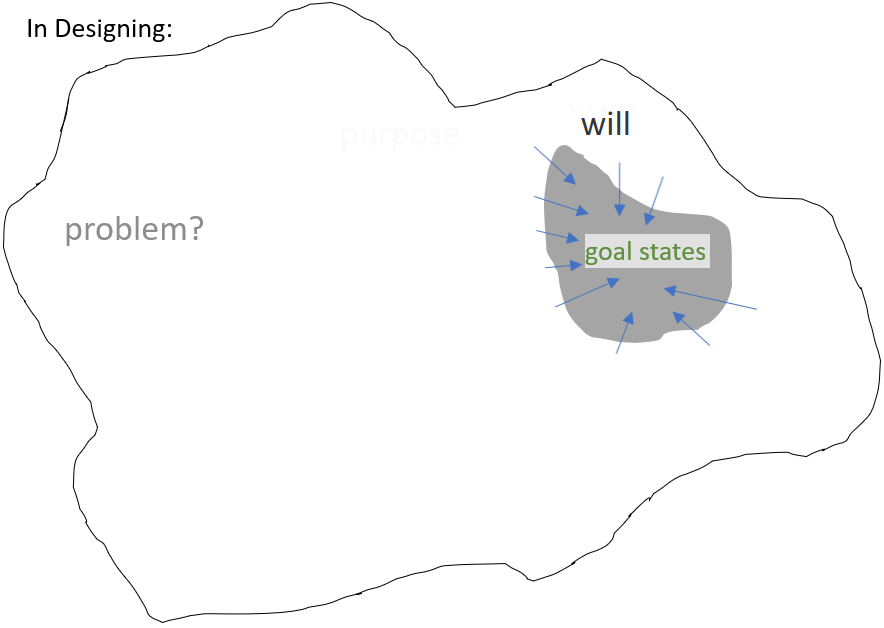}
	\caption{Designing approach}
	\label{fig:designing}
\end{figure}

While in problem-solving, the will was generated from a problem, i.e. growing from an initial state, in designing it is the opposite. Here, instead, it is growing from the final state(s), searching for the best state to start the full solution from. It is like creating a story backward: starting from the end, to reach some beginning (Fig~\ref{fig:designing}).

In designing there is a goal and a will to go there, but no specification of the problem or the initial states.
So it is an iterative process, starting from searching for a problem to reach the goal, then continuing with a specific will connecting the problem with the goal, resulting with a problem to solve. 

\subsubsection{Summary of Problem-Solving and Designing} \label{sec:summary}

One can spot a duality here also.
On the one hand, problem-solving is an analytical perspective, looking for a resolution or finishing a problem, by usually a systemic view, breaking it to parts and then looking for some appropriate solution, that serves as a better state than the one we started with (problem state).
On the other hand, designing is based on a holistic perspective, where instead of finding a fast/analytical resolution to a problem (“to make everyone happy and go on with our lives”) it is about empathy/consideration, i.e. it is about looking for the roots of the problem, and not just shutting it down quickly. It takes the opposite approach: instead of reducing the problem, it tries to track its sources and thus solving the causes that generated this problem state/situation. By doing so, it searches for a better problem to solve, which solves multiple other problems.
Either way, will 
is either getting somewhere (goal) or getting out of something (problem).

\newpage

\subsection{Implementation} \label{sec:Implementation}

\subsubsection{Combining Will and Modeling} \label{sec:combination}

\begin{figure}[H]
	\centering
	\includegraphics[width=0.95\textwidth]{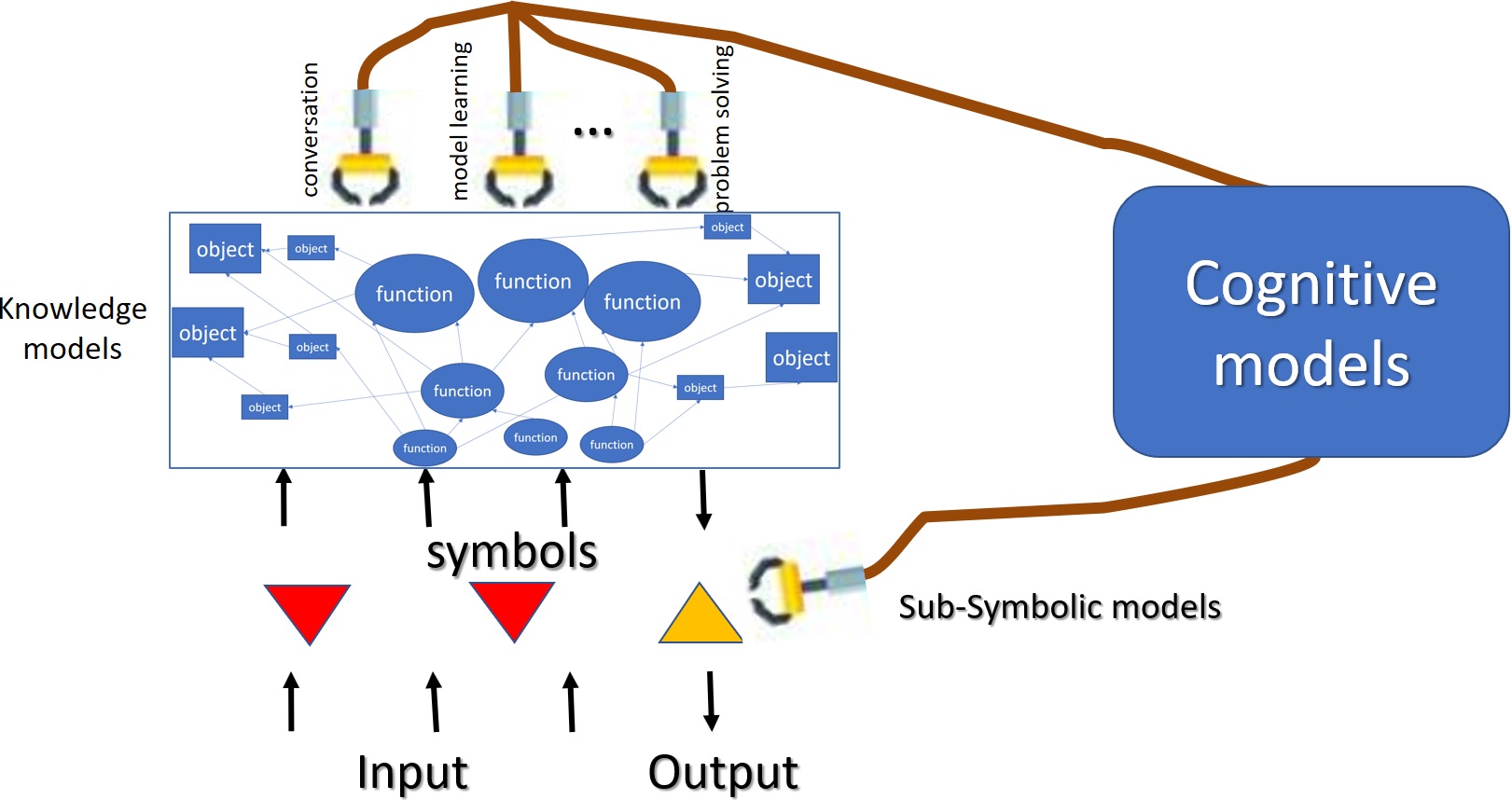}
	\caption{Separating cognitive and knowledge models}
	\label{fig:Modelling_machines3}
\end{figure}

The final question, is how all the discussed above, i.e. in \ref{sec:problem_solving}, and in \ref{sec:AKREM}, i.e. will-related topics, are combined with the operational model. There are a few options, as discussed in \ref{sec:future_work}, but we emphasize one of them here.
One option, is to have one main supervising model, that depending on the category of the situation, assigns the proper model to handle it. E.g. model for problem-solving, for learning, and for story message (where it connects separate events sequentially).
Conversation for example is about taking turns, waiting till me/other side finished, recognizing our models, etc. Perceiving fictional information is treated differently than factual information, and so on.
In conclusion, this option derived since problem-solving and alike are very complex models, which is why the suggestion to separate them from the knowledge models. But it extends further - perhaps there is separation of model representation. May be some models can be represented as operational classes, but others cannot. These others could be not interpretable nor can be explained by the agent, since they are in the background of thinking itself, thus they are “hidden” or implicit. 
See Fig.~\ref{fig:Modelling_machines3}.

\newpage

\subsubsection{Multi-scale Consolidation in Model Learning} \label{sec:consolidation_in_learning}

\begin{figure}[H]%
	\centering
	\subfigure[Nested DNN structure]{%
		\label{fig:a}%
		\includegraphics[width=0.99\textwidth]{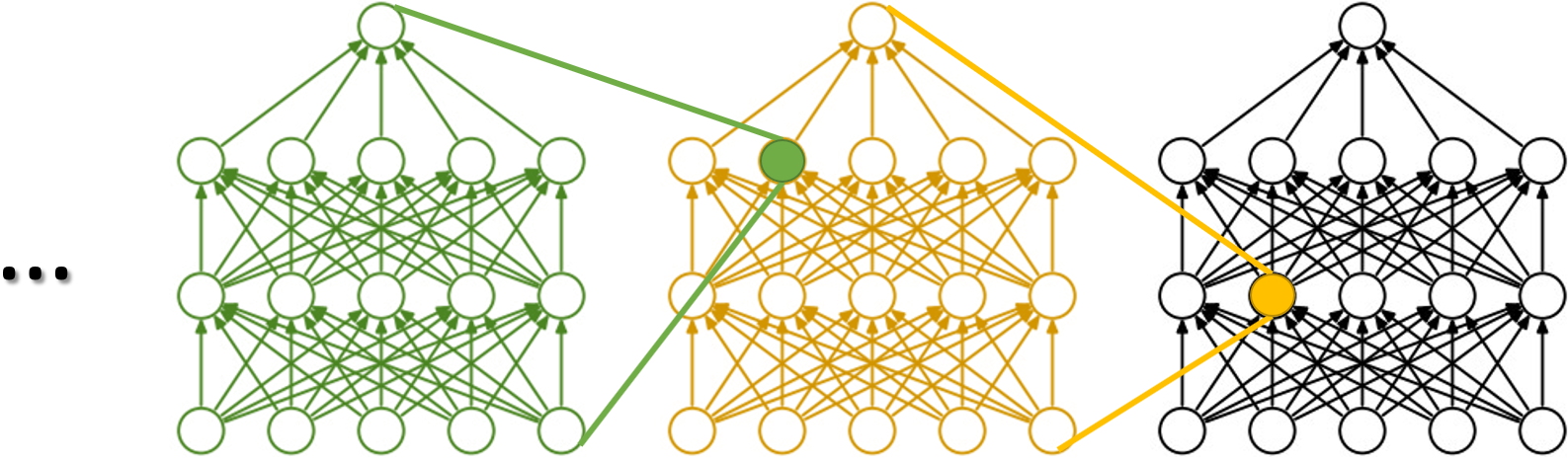}}%
	\hfill
	\subfigure[Initial Nested DNN]{%
		\label{fig:b}%
		\includegraphics[width=0.49\textwidth]{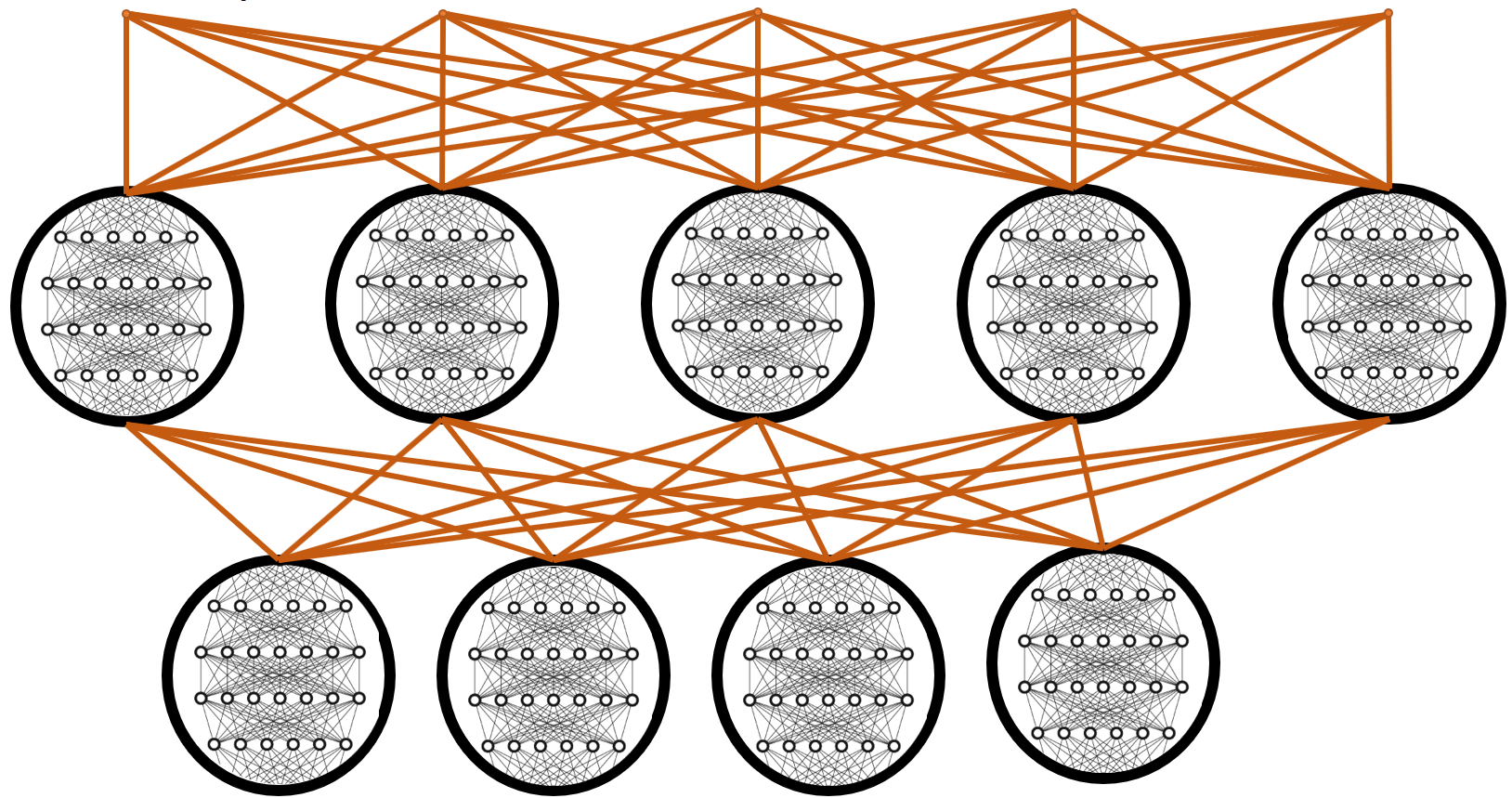}}%
	\hfill
	\subfigure[Learned Nested DNN]{%
		\label{fig:b}%
		\includegraphics[width=0.49\textwidth]{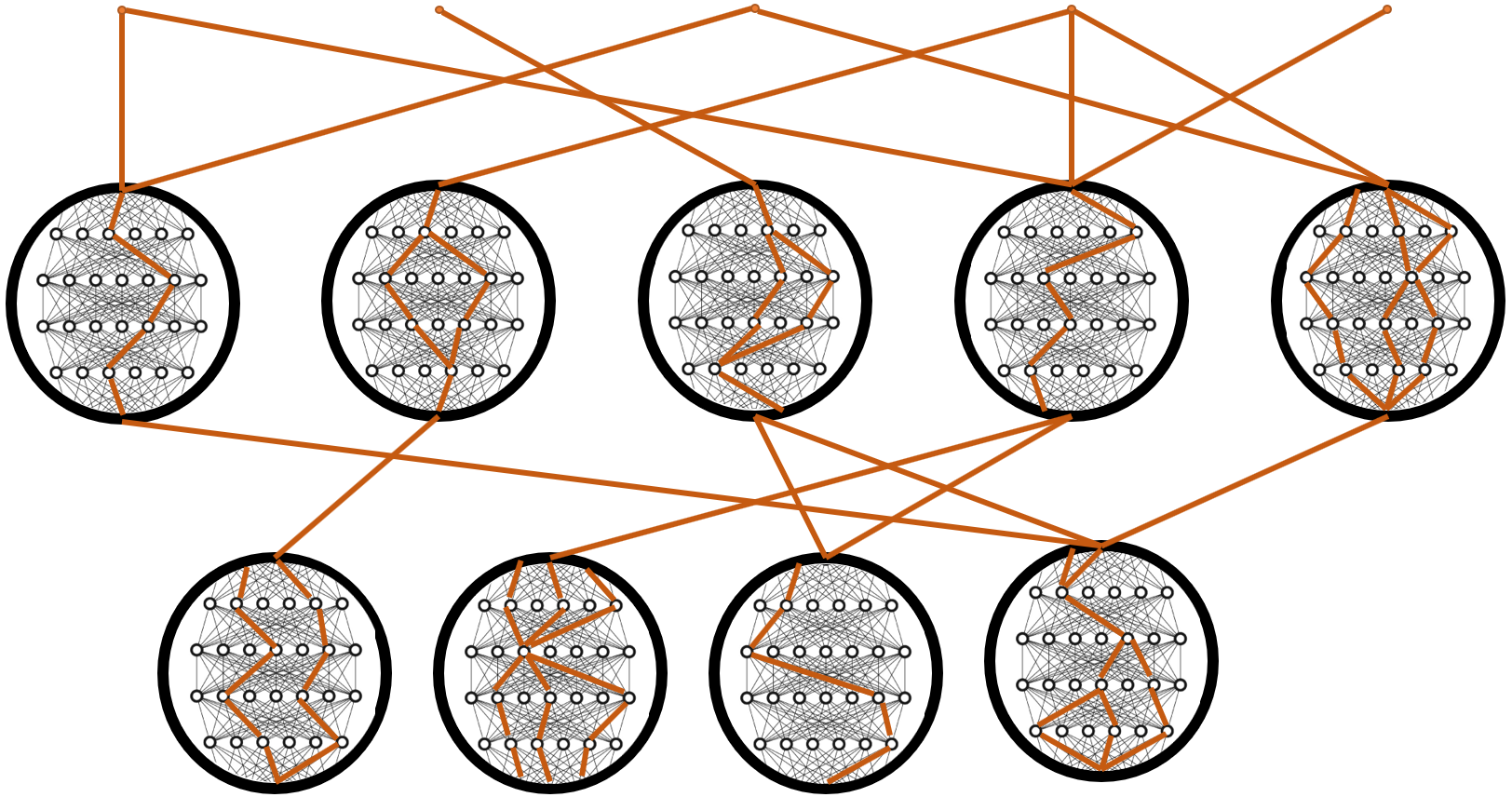}}%
	\caption{Nested DNNs for model learning}
	\label{fig:Multiple_Consolidation}
\end{figure}

Next question, is how models can be learned separately? One solution is by assuming some initial network of unlearned models, see Fig.~\ref{fig:Multiple_Consolidation}(b). But before that we assume consolidation in multiple scales, as if there is consolidation also in scaling (discrete amount of them). It can be encountered through many phenomenon in nature, e.g. in the universe (consolidation into stars/solar-systems and galaxies), in fractals (such as snow-flakes), and in other recursive structures. See for example the nested structure in Fig.~\ref{fig:Multiple_Consolidation}(a), and in the transition from Fig.~\ref{fig:Multiple_Consolidation}(b) to Fig.~\ref{fig:Multiple_Consolidation}(c), consolidation in multiple levels, both in micro (within models), and in macro (between models). 
We see that modeling or reorganization of inner elements, is occuring at many levels of models, i.e. from the basic models to the most complex ones.

\newpage

\subsection{Attention} \label{sec:attention}

After assuming a bunch of unlearned models, Fig.~\ref{fig:Multiple_Consolidation}(b), we can assume that will, acting like a flashlight or a beacon, produce consistent attention (over time) to learn/attend each model (or several of them) separately.

\begin{figure}[H]%
	\centering
	\subfigure[Attending specific model1]{%
		\label{fig:a}%
		\includegraphics[width=0.49\textwidth]{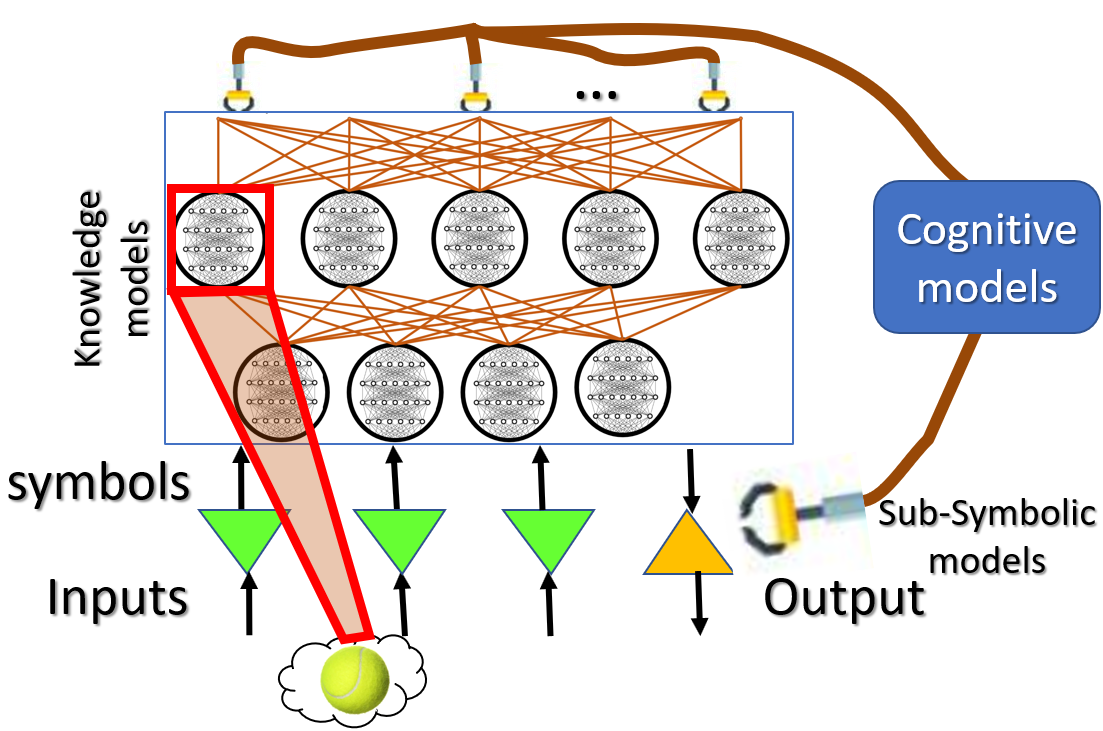}}%
	\hfill
	\subfigure[Attending specific model2]{%
		\label{fig:b}%
		\includegraphics[width=0.49\textwidth]{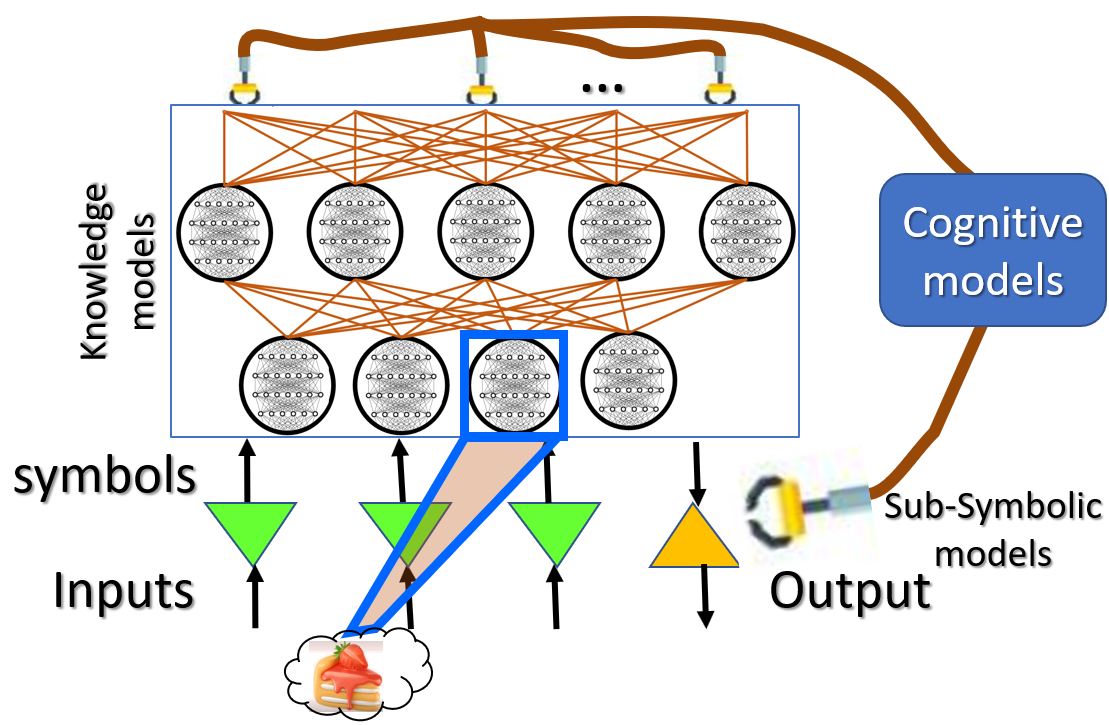}}%
	\caption{Local model learning via attention}
	\label{fig:attnetive_model_learning}
\end{figure}

This explains why an infant is usually very focused over his toys (e.g. a ball), and tracking them is essential for this process. This effect stick till adulthood, also in the process of using the cognitive model (i.e. after the learning stage). It is the need to be attentive only to a limited set of models ($7\pm 2$ items in WM).
See examples in Fig.~\ref{fig:attnetive_model_learning}.

Next question is how this will is applied in story telling/hearing and in problem solving? 
For example, in my presentation, the problem-solving 3-layered slide actually showing these beams, searching for solutions!

We can see it in the following Fig.~\ref{fig:attnetive_model_learning2}.

\begin{figure}[H]
	\centering
	\includegraphics[width=0.72\textwidth]{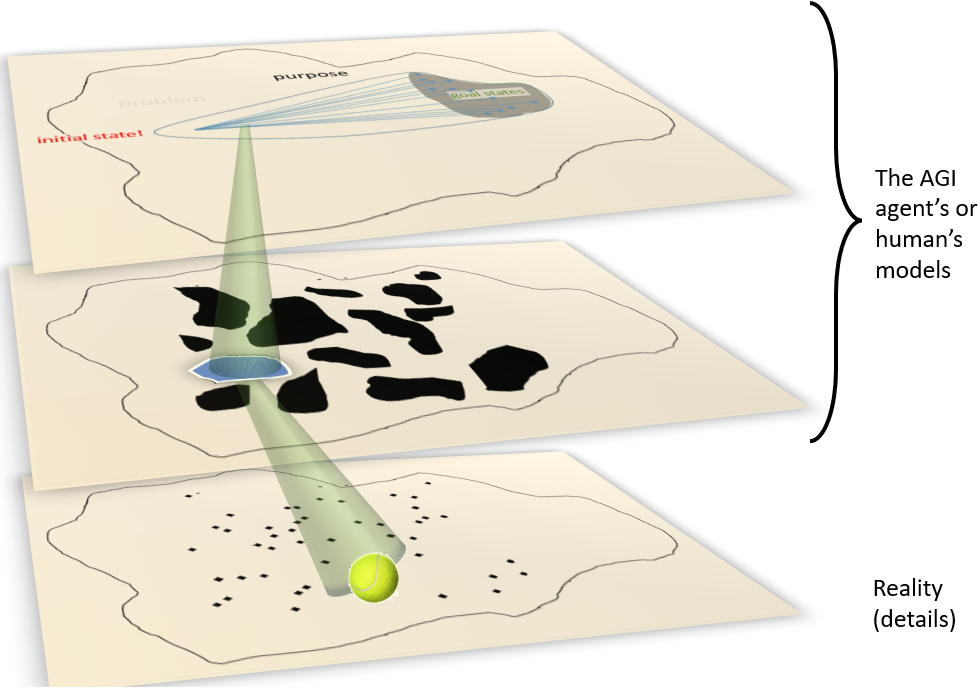}
	\caption{Learning reframed as problem-solving task}
	\label{fig:attnetive_model_learning2}
\end{figure}

We can say that learning, is making-sense type of problem-solving. So again, there is will, coming from the top, like a projector, focusing on one/few models, while tracking it in the real world. More accurately, we should have changed the top will, not as a purpose to reach but as the point-wise will at the start of a problem.

Moreover, we could say that the hierarchy is abstraction, as claimed earlier, and will is actually only on the top but "shining" directly on the models. Then, we could say that during waking hours, an infant is gathering instances of its current models, e.g. a ball, and at sleeping, he uses these instances to train his models, for the purpose of making sense. The waking hours do not do it, they can only perform cognitive operations, which is actions in these models. So first, he tries to figure out different models, then he tries to model them also in time, thus able eventually to track them, which is a validation of the correctness of his "ball" model for example. Because the final test of his model is prediction, hence temporal modeling is what enables prediction, or more specifically forecasting (prediction in time).

Note, that attention to a few models also implies that just as humans, AGI agent need not to understand and model everything, but only what it is focused on or interested with.
Also, there is the idea of bidirectional attention, which is bottom-up (external) verse top-down (its own will), and describes the competition between having a (strong) will to be highly influenced by the outside. In AGI's case, it should be mostly navigated by external guidance, if will is not engineered into it.

In addition, attention can have different "focal length", like the theory of vision, having small pinhole perception at lower levels, and a bigger one at higher ones. Meaning, the ability to sometimes see small details and sometimes see the big picture. In model attention it is the same: we can both have low-level more detailed attention on smaller models, upto a high-level attention for more general or composite models. In comparison to classical object detection in computer vision, high-level concepts use only the higher-level features for the classification task, but more generally there is no reason not to be attentive to low-level features whenever is needed.

Finally, attention in our perspective is very similar to the attention in DL, only without regulation. Meaning, without considering the ideal state of consolidating into symbolic reasoning of models and operations and most importantly allowing for dynamic abstraction. However, DL's attention is similar in that it too allow for multiple implicit functions in a given learning NN, since it react differently depending on the input. In other words, the DNN can be regarded as a group of undeclared models/functions, generated by attention units, thus implicitly implement compositionality and reusability.


\newpage


\bibliography{../Shimon_paper02,../Shimon5}
\end{document}